\documentclass[
 reprint,
 amsmath,amssymb,
 aps,
]{revtex4-1}

\usepackage{graphicx}	% Include figure files
\usepackage{dcolumn}	% Align table columns on decimal point
\usepackage{bm}		% bold math
\usepackage{hyperref}	% add hypertext capabilities
\usepackage{comment}
\usepackage{color}
\usepackage[caption=false]{subfig}
\usepackage{commath}
\usepackage[export]{adjustbox}

\DeclareMathOperator{\erf}{erf}

\usepackage{amsfonts}
\usepackage{amssymb}

\begin{document}
%\preprint{APS/123-QED}

\title{Learning Direct and Inverse Transmission Matrices} % Force line breaks with \\

\author{Daniele Ancora}
\email{daniele.ancora@nanotec.cnr.it}
\affiliation{CNR-NANOTEC, Institute of Nanotechnology, Rome Unit, Soft and Living Matter Lab., Piazzale Aldo Moro, 5, 00186, Rome, Italy}
\author{Luca Leuzzi}%
\affiliation{CNR-NANOTEC, Institute of Nanotechnology, Rome Unit, Soft and Living Matter Lab., Piazzale Aldo Moro, 5, 00186, Rome, Italy}
\affiliation{Physics Department, Sapienza University,  Piazzale Aldo Moro, 5, 00186, Rome, Italy}


\date{\today}% It is always \today, today,

\begin{abstract} % IT MUST BE WITHIN 600 CHARACTERS
%Linear problems are largely used in all the scientific fields, ranging from ... till the description of the signal transmission in disordered system. Among its generality, the non-trivial problem of the transmission matrix recovery of random media is one of the most stimulating challenges in the field of biomedical imaging. Its knowledge would turn any opaque layer into a normal optical tool, capable to focus or transmit an image through disorder, turning the scattering into a beneficial feature exploitable to enhance resolution. Many studies have pursued the solution of this riddle, achieving light-focusing control or reconstructing images behind complex media. In the present work, we investigate how statistical inference could help the calculation of the transmission matrix in a complicated light-scrambling environment. We convert a generic linear input-output transmission problem into a statistical mechanical formulation and, based on pseudolikelihood maximization algotihms, we learn the coupling matrix via randomly sampling intensity realizations. With this work we propose a new statistical framework, bridging linear regression and thermodynamical approaches to uncover insights from the scattering problem.

Linear problems appear in a variety of disciplines and their application for the transmission matrix recovery is one of the most stimulating challenges in biomedical imaging. Its knowledge turns any random  media into an optical tool that can focus or transmit an image through disorder. Here,  converting an input-output problem into a statistical mechanical formulation, we investigate how inference protocols can learn the transmission couplings by pseudolikelihood maximization. Bridging linear regression and thermodynamics let us propose an innovative framework to pursue the solution of the scattering-riddle.

%\begin{description}
%\end{description}
\end{abstract}

		% PACS, the Physics and Astronomy Classification Scheme.
 	% Use showkeys class option if keyword display desired
\maketitle

%\tableofcontents

%%%%%%%%%%%%%%%% SECTION
%\section{\label{sec:introduction}Introduction}
A major interest in biomedical imaging is the comprehension of the light scattering through disordered media: many recent studies have achieved light-focusing and image reconstruction even through complex biological tissues \cite{bertolotti2012non, vellekoop2010exploiting}. The deterministic nature of the scattering event suggests that turbid devices could be treated as a normal optical tool: the memory effect principle \cite{Freund1988} states that tilting the illumination wavefront turns into a translation of the scrambled intensity pattern. This makes the turbid layer acting as an autocorrelation-lens \cite{Freund1990}, possible to be used for focusing or imaging through -or even behind- opaque walls. More generally, any small variation of the input wavefront results into a small variation of the output in a deterministic and continuous fashion, thus a transmission matrix approach was proposed to describe the process \cite{Popoff2010}. In analogy with classical optics, the transmission matrix would contain the -yet complicated- rules on how the device acts on a given input, transporting it into a disordered output via a linear combination. Although few important studies were done \cite{Popoff2011, Yoon2015, Carpenter2014}, measuring such matrix is one of the greatest challenges in disordered photonics \cite{Wiersma2013} and could give new insights on the scattering process. Among a number of possible applications, the main interest of the biomedical imaging community is toward its measurement in a disordered multi-mode fiber transmission \cite{Flaes2018}, that would open up their usage against the more fragile and expensive single-mode bundle fibers counterpart in endoscopic devices. 
Put in simple words, the problem is that whatever we send in the input appears totally randomized at the output, due to the complex photon paths permitted by the complex media. An established option, at the moment, is the method provided by Popoff et al. \cite{Popoff2011} that relies on the usage of the Hadamard basis for the input to calculate the transmission matrix based on output observations. Since it relies on output sampling on a given input basis, this method is quite specific and, in practice, one needs to accomplish $\mathbb T$-matrix inversion before being able to effectively focus or image through disorder. 
Our study fits in this scenario: we investigate how statistical inference could offer a novel way to learn the transmission matrix $\mathbb T$ of a complex media, unbounded from any basis and even free from matrix inversion. In the following, we will introduce the mathematical model, inspired by a spin-like thermodynamic description of an interacting input/output Hamiltonian. The analogy with spin-glass theory let us borrow a number of statistical tools for the inference of the coupling parameters, tightly linked with the unknown $\mathbb T$. In particular, we make use of a pseudolikelihood maximization approach coupled with a progressive parameter decimation scheme \cite{Decelle2014}. On the other hand, for very sparse matrices one might consider the recent activation technique \cite{rocchi2019}. The model, in these terms, is scalable to any number of parallel implementations, avoiding exponential complexity for the solution of the inverse problem. At the current stage, we leave our approach general in terms of applicability, promoting it also as a thermodynamic alternative to any linear regression problem \cite{yan2009linear}.

%%%%%%%%%%%%%%%% SECTION
%\section{\label{sec:mathematical}Mathematical Framework}
We start considering a linear transmission problem involving a two-edge input-output system (from now on I/O). In these terms, the disordered medium acts as a linear light scrambler, connecting the input $I^{in}$ (described by the index $\alpha = 1,..., N/2$) to the output pattern $I^{out}$ (index $\gamma = 1,..., N/2$) via an intensity transmission matrix $\mathbb T$ of the form given by:
\begin{equation} \label{eq:transmission}
I^{out}_\gamma = \sum_{\alpha=1}^{N/2} T_{\gamma\alpha}I_\alpha^{in} + \sigma_\gamma \epsilon_\gamma.
\end{equation}
\begin{figure}[h]
\centerline{\includegraphics[width=1.00\columnwidth]{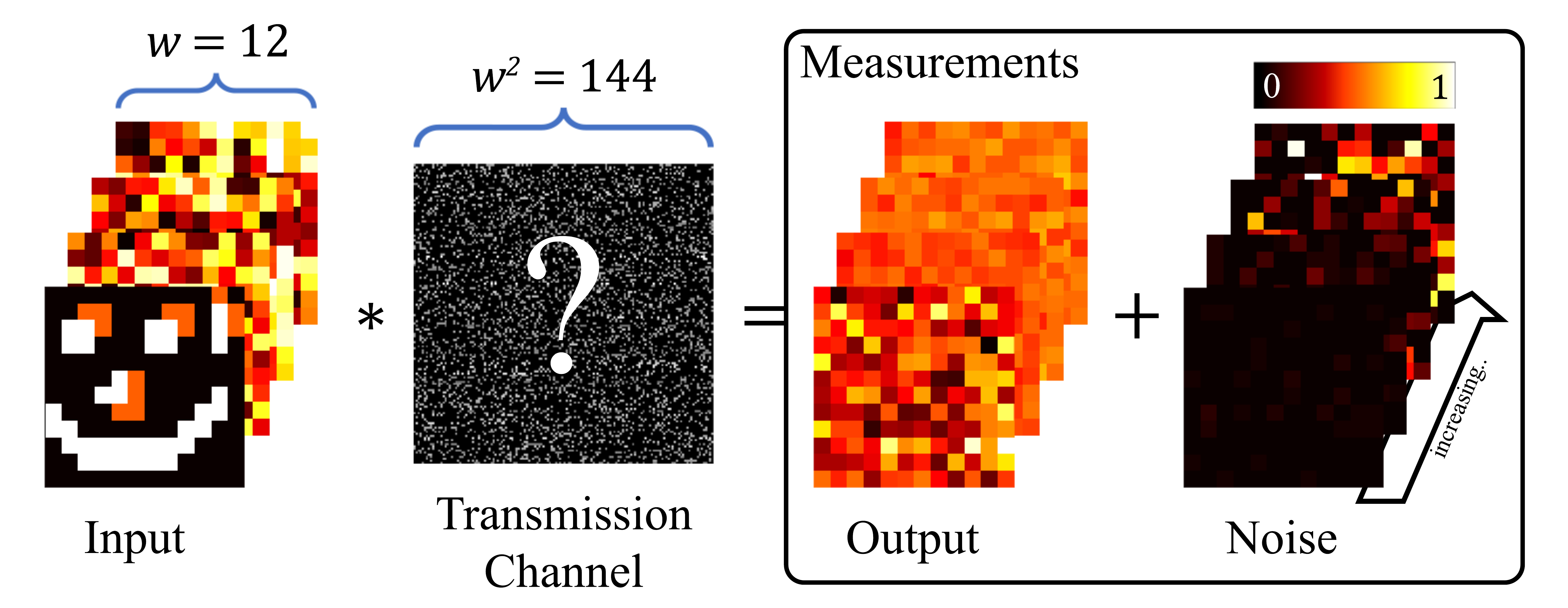}}
\caption{Schematics of the problem. We send a given input set of $w\times w$ pixels through an unknown transmission channel, measuring the noisy output on the other edge. The output is not trivially connected to the input, resulting in the production of a seemingly informationless speckle patter. We want to learn the parameters $\{T_{\gamma\alpha}\}$ of the channel via a random sampling statistical inference approach.}
\label{fig:scheme}
\end{figure}
In Eq. (\ref{eq:transmission}), the last factor is a noise term given by the product of $\epsilon$, a vector containing normally distributed random numbers. The term $\sigma_\gamma$ rescales the mean square displacement in the channel $\gamma$. Let us initially  set $\sigma_\gamma=0$, $\forall \gamma$.
This is a simplified optical model, due to the fact that we are connecting the intensities measured from both fiber's ends rather than treating more appropriately complex electromagnetic fields \cite{Popoff2010,Tyagi2016,Marruzzo2017a}. With such assumption we are neglecting the phases that cannot be directly measured with current technology, looking only at the modulus squared of the field amplitude at each site. This method would work rigorously with incoherent sources and serves to introduce the statistical framework.
%In a further implementation we will extend it to complex field analysis, treating explicitly the phases or integrating them out, reproducing a more accurate inference model for coherent sources.
In this representation $\mathbb T$ is not exactly the electromagnetic transmission matrix of the fiber, but rather a matrix connecting the intensities between the fiber's ends: however, we will refer to it as an effective-$\mathbb T$, leaving the discussion of the more realistic case to further studies. We stress, however, that the present model has the advantage to be easily applicable to any generic experimental I/O pattern.

The recovery of the matrix $\mathbb T$ corresponds to find the solution of all equations  (\ref{eq:transmission}), such that:
\begin{equation} \label{eq:delta}
\int \prod_{\gamma= 1}^{N/2} dI_\gamma^{out} \delta \left( I^{out}_\gamma - \sum_{\alpha=1}^{N/2} T_{\gamma\alpha}I_\alpha^{in} \right) = 1.
\end{equation}

Assuming Gaussian noise-induced uncertainty to the solutions of these equations corresponds to approximate the $\delta$-functions in (\ref{eq:delta}) with Gaussian functions having their variance vanishing to zero, so that we can write Eq.  (\ref{eq:delta}) as
\begin{equation} \label{eq:delta_lim}
%\lim_{\Delta\rightarrow 0} \frac{1}{\sqrt{2\pi\Delta^2}} \int \prod_\gamma dI_\gamma^{out} e^{\left( -\frac{\left(I^{out}_\gamma - \sum_\alpha T_{\gamma\alpha}I_\alpha^{in}\right)^2}{2\Delta^2}\right)}= 1
%\end{equation}
\lim_{\Delta\rightarrow 0}  \int \prod_{\gamma=1}^{N/2} dI_\gamma^{\rm out} \  \frac{e^{  -\frac{1}{2\Delta^2}  \left(I^{\rm out}_\gamma - \sum_\alpha T_{\gamma\alpha}I_\alpha^{\rm in}\right)^2}}{\sqrt{2\pi\Delta^2}}= 1.
\end{equation}

The mean square displacement $\Delta$ corresponds to the uncertainty term $\sigma_\gamma$ in (\ref{eq:transmission}). For the sake of simplicity we initially consider the same noise for each channel.    Let us call $\mathcal{H}$ the squared argument on the exponent, that can be readily expressed as:
\begin{align}
\mathcal{H} &= \sum_\gamma \left(I^{out}_\gamma - \sum_\alpha T_{\gamma\alpha}I_\alpha^{in}\right)^2 = \nonumber \\
&= \sum_\gamma (I_\gamma^{out})^2 - 2\sum_{\gamma, \alpha} I_\gamma^{out} T_{\gamma\alpha} I_\alpha^{in} + \sum_{\gamma, \alpha,\alpha'} T_{\gamma\alpha}T_{\gamma\alpha'}I_\alpha^{in}I_{\alpha'}^{in} \nonumber \\
&\equiv \sum_{i,j} I_i J_{ij} I_j.\label{eq:hamiltonian}
\end{align}
To reach the matrix formulation (\ref{eq:hamiltonian}) we definined a generic vector $\bm I =\{ I_{\gamma}^{in},I_{\alpha}^{out} \}$ of length $N$ obtained concatenating both the elements of the input and output fields, where both indexes $i,j = \{ \gamma, \alpha \}$ span the concatenated input and output indexes. 
%Thus, in vector notation $\mathcal{H} = \bm I^T \mathbb J \bm I $. 
Explicitly, the interaction matrix $\mathbb J$ has the form of a tensor that contains the transmission matrix and its conjugate as expressed in:
\begin{equation} \label{eq:generalizedJ}
\mathbb J=
\begin{pmatrix}
    -\mathbb{U}	& +2\mathbb T^\dagger \\
    +2\mathbb T	& -\mathbb{I} \\
\end{pmatrix}
\end{equation}

\noindent where $\mathbb{U}=\mathbb{T}^\dagger \mathbb T$ is the input self-coupling Gramian matrix, $\mathbb T$ is the transmission matrix defined in Eq. \eqref{eq:transmission} and $\mathbb{I}$ is the identity matrix. $\mathbb J$ is  the generalized coupling matrix that we will infer with a statistical approach. In this framework the system can be seen as a generalized spin-like I/O model described by the Hamiltonian $\mathcal{H}= \bm I^T \mathbb J \bm I $.

Giving a statistical interpretation to the Boltzman factor $e^{-\beta\mathcal{H}}$ in \eqref{eq:delta_lim}, the probability of a coupled input-output realization $\bm I$ with a given Hamiltonian disaplying coupling constants $J_{ij}$ is equal to:
\begin{equation} \label{eq:probability}
P(\bm I | \mathbb J) = \frac{1}{\mathcal{Z}(\mathbb J,\bm I)} \exp\left\{-\beta \sum^{1,N}_{ij} I_i J_{ij} I_j \right\}
\end{equation}
where the partition function of the system is:
\begin{align} \label{eq:partition_function}
\mathcal{Z}(\mathbb J,\bm I) = \int  \prod_{k=1}^N dI_k  \  e^{-\beta \mathcal{H}(\mathbb J,\bm I)}.
\end{align}
\noindent with the definition $\beta = (2\Delta^2)^{-1}$, to be interpreted as the inverse temperature of the I/O model, relative to the effective noise term in the measurements. Given the probability, we can write the (log-)likelihood  that a system with couplings $\{J_{ij}\}$ yields 
the experimental I/O realizations $\bm I$:
\begin{equation}
\mathcal{L} =  \ln \big[ P(\bm I | \mathbb J) \big].
\end{equation}
The set of couplings $J_{ij}$ that maximizes the likelihood are the ones that most likely represent the observed model $\mathcal{H}(\mathbb J,\bm I)$ given all the possible realizations of $\bm I$.

%%%% subsection
%\subsection{\label{ssec:pslIO}Pseudolikelihood of the I/O model}
Unfortunately $\mathcal{L}$ is very difficult to maximize due to an exponential complexity growth as a function of the number of parameters to be inferred. It is convenient in this case, to consider a less computational expensive approach by introducing the log-pseudolikelihood function $\mathcal{PL}$ \cite{Ravikumar10,Aurell12}. We proceed fixing all intensities except the $i$-th, and we write the conditioned probability of a  value of $I_i$ given the values of all the other pixels $\bm I_{\backslash i}$ \footnote{We drop the depedence on $\mathbb J$ to shorten the notation}: 
\begin{equation}
P(I_{i}|\bm I_{\backslash i}) = \frac{P(\{I_i, \bm I_{\backslash i}\})}{P(\bm I_{\backslash i})}.
\end{equation}

The function $\mathcal H$ is separable in $N$ independent partial functions $\mathcal{H}_i$ as
\begin{equation}
\mathcal{H} = \sum_{i=1}^N \left[{I_i \sum_{j \neq i} J_{ij} I_j  + I_{i}^2 J_{ii}}\right] = \sum_{i=1}^N \mathcal{H}_i 
\label{eq:factor}
\end{equation}
We, further, define  $A_i \equiv -\beta_i J_{ii}$ and ${B_i \equiv \beta_i \sum_{j \neq i} J_{ij}I_j }$ such that
\begin{equation}
\beta_i \mathcal{H}_i = I_iB_i[\bm I_{\backslash i}] - I_i^2 A_i.
\end{equation}
 We notice that we leave the parameters $\beta_i$ free to variate for each pseudo-likelihood, to quantify the noise at each channel independently. In fact, thermodinamically, this factor could be seen as an inverse channel temperature for $i=N/2+1, \ldots, N$ (when it is expected $J_{ii}=1$).
By definition, the $A_i\geq 0$, $\forall i$, guaranteeing the convergence of  the partition function.
Using Eq. \eqref{eq:factor} we write the pseudolikelihood of the variable $i$ as
\begin{eqnarray}
P(I_{i}|\bm I_{\backslash i}) &=&  \frac{e^{\beta_i \mathcal{H}_i(I_i)}}{\mathcal{Z}_i[\bm I_{\backslash i}]}
 \label{eq:partitionfunction}
 \\
\mathcal{Z}_i[\bm I_{\backslash i}]&\equiv&  \int dI_i \ e^{\beta_i \mathcal{H}_i(I_i)} = \int dx \  e^{-A_i x^2+B_i[\bm I_{\backslash i}] x} 
\nonumber 
 \end{eqnarray}

% % % % %  FIN QUI 
\begin{comment}
Now that we have separated the probability distribution, it is possible to write explicitly their form in each site $i$:
\begin{equation}
P(I_{i}|\bm I_{\backslash i}) = \frac{e^{I_i {\sum_{j \neq i} J_{ij} I_j  + I_{i}^2 J_{ii}}}}{\int d I_i e^{I_i {\sum_{j\neq i} J_{ij} I_j  + I_{i}^2 J_{ii}}}} = \frac{e^{\beta_i \mathcal{H}_i}}{\int d I_i e^{\beta_i \mathcal{H}_i}}
\end{equation}
and with this equation it is possible to define
\end{comment}
\noindent
and the log-pseudolikelihood per  element  $i$ is:
\begin{equation} \label{eq:psl}
\mathcal{L}_i = \ln P(I_{i}|\bm I_{\backslash i})  = \beta_i \mathcal{H}_i  - \ln \mathcal{Z}_i.
\end{equation}

\begin{comment}This formulation was proven to converge to the exact likelihood function for a high number of measurements:
\begin{align}
\mathcal{L} &= \ln P(\bm I) \simeq \ln \prod_{i=1}^{N} P(I_{i}|\bm I_{\backslash i}) = \sum_{i=1}^{N} \ln P(I_{i}|\bm I_{\backslash i})  =  \nonumber \\
&=  \sum_{i=1}^{N} \mathcal{L}_i = \mathcal{PL}.
\label{eq:totalPSL}
\end{align}
\end{comment}

Two approaches are available at this stage: we minimize all the $\mathcal{L}_i$ in function of the coupling matrix $\mathbb J$ using some regularization \citep{Tyagi2016}, or we minimize their sum 
\begin{equation}
\mathcal{PL}\equiv \sum_{i=1}^N \mathcal{L}_i 
\label{eq:totalPSL}
\end{equation}
that we refer to as total log-pseudolikelihood function. The latter is commonly followed by a decimation procedure \cite{Decelle2014}, in which we recoursively set to zero small couplings until the total pseudolikelihood is not substantially affected by such  change.
%% From now on we drop the log- prefix without loosing specificity, referring directly to likelihood and pseudolikelihood functions.

%%%% subsection
%\subsection{\label{ssec:partitionfunction}Which partition function?}
%{\color{red}{[\bf Questa parte sotto potremmo toglierla indicando solo quale scelta adottiamo per di risultati che mostriamo nella lettera e riferendoci all'articolo lungo per il resto.]}}
%\vskip 1 cm

It is obvious that $\mathcal{L}_i$ strictly depends upon the integration extremes of the partition function. The choice of the integral extremes in Eq. (\ref{eq:partitionfunction}) has to take into account all the possible intensities allowed by the system. Although the intensity treated are always in a limited range, we found that the general choice of integrating along the whole real axis works for the intensity ranges tested. Thus, the undefined integral (\ref{eq:partitionfunction}) can be calculated between $(-\infty,\infty)$:
\begin{align}
\mathcal{Z}_i&= \sqrt{\frac{\pi}{4A}} e^{\frac{B^2}{4A}} \erf {\left( \frac{-B+2A I_i}{\sqrt{4A}} \right)} \bigg|_{-\infty}^{+\infty}
\\ &= 2 \sqrt{\frac{\pi}{4A}} e^{\frac{B^2}{4A}}.
\nonumber
\end{align}
This simple operational choice for the extremes turns out to be the most effective against more strict integration ranges (see Ref. \cite{Ancora19b} for details), leading to:
%. In this framework, the $\mathcal{L}_i$ can be explicitly written as:
\begin{align}
\mathcal{L}_i = I_i B_i[\bm I_{\backslash i}] - I_i^2A_i - \frac{1}{2} \ln \left( \frac{\pi}{4A_i} \right) - \ln 2 \ . 
\end{align}

\begin{figure}[b!]
\centerline{ \adjincludegraphics[width=8cm,trim={0 {.05\height} 0 {.07\height}},clip]{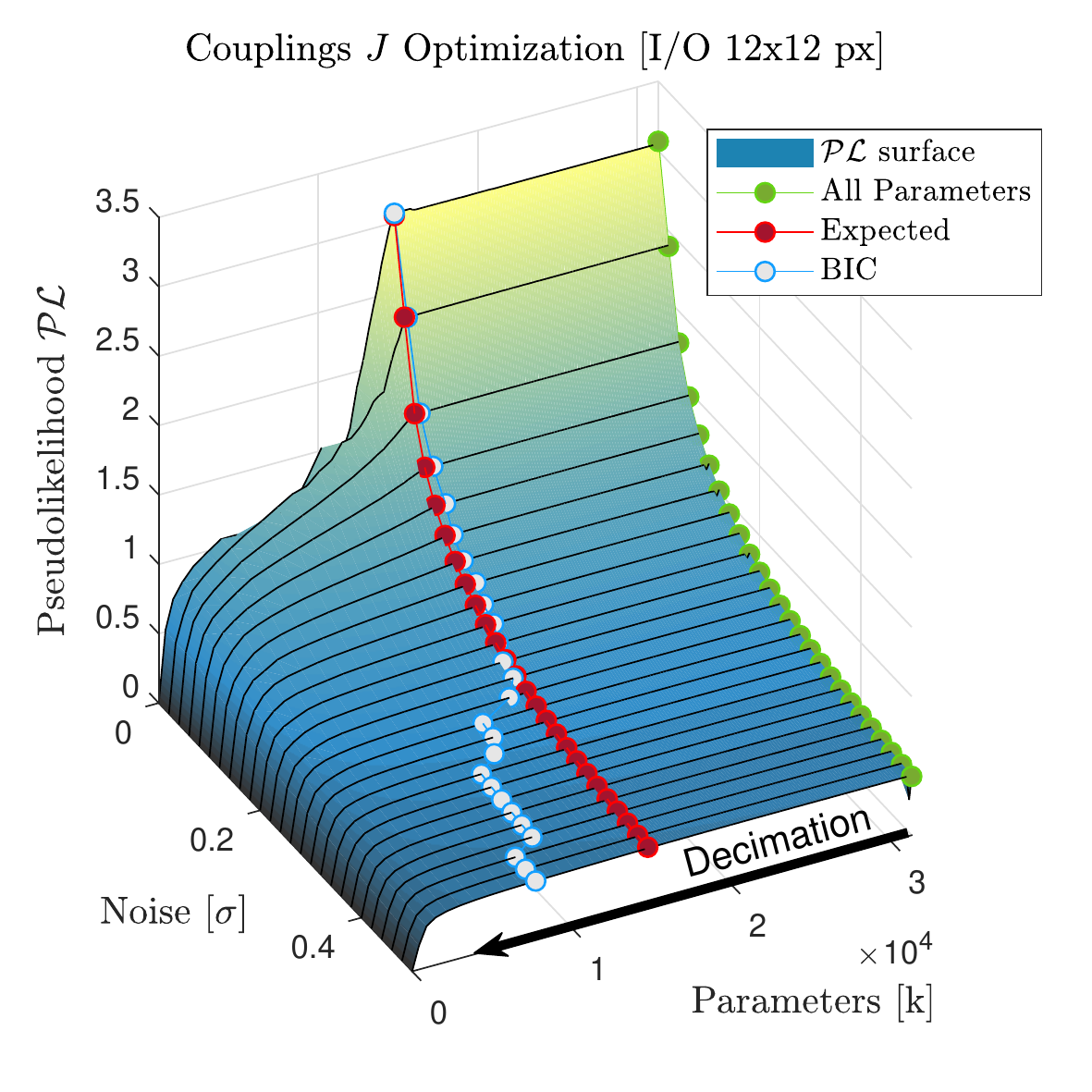}}
%\centerline{\includegraphics[width=8cm]{Fig1_directT_official}}
\caption{Maximized pseudolikelihood $\mathcal{PL}$ surface as a function of the number of parameters and noise. The noisier the channel, the smoother the transition between an optimal maximum $\mathcal{PL}$ and the value at which it drops due to underfitting.}
\label{fig:PSLsurf}
\end{figure}

\begin{figure}[t!]
\centerline{\includegraphics[width=8cm]{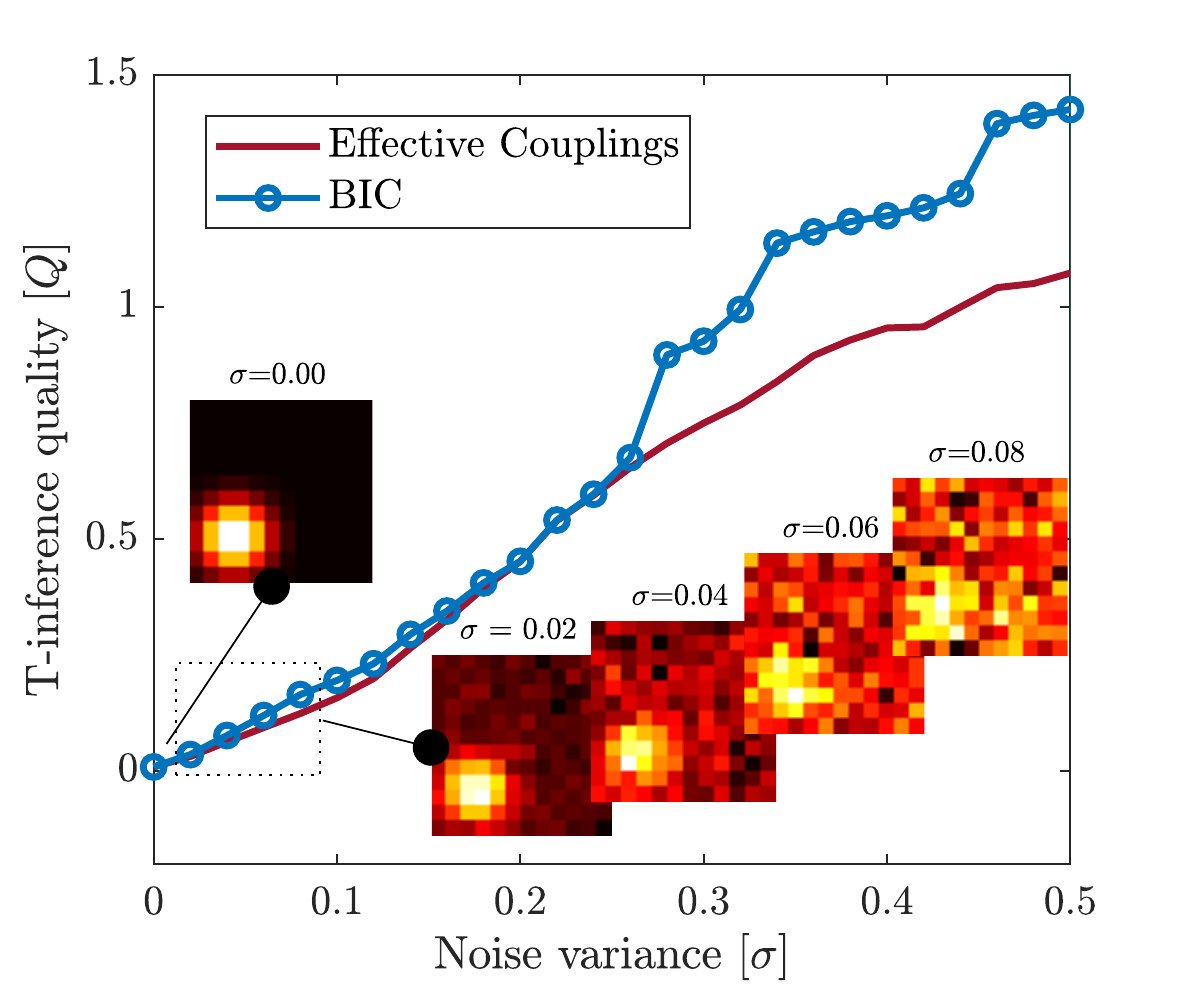}}
\caption{Transmission parameters error in function of the output noise. The red curve is knowing the number of active channels, the blue is using BIC selection. The images in the inset show an ideal focusing experiment using the recovered $T$-information. }
\label{fig:errorFocusing}
\end{figure}

%%%%%%%%%%%%%%%% SECTION
%\section{\label{sec:methods}Numerical Methods and Results}
By definition,  $-\mathcal{L}_i$ are convex functions that can be minimized using a quasi-Newton method. 
%%%Thermodynamically, it is equivalent to maximizing the entropy of the model in function of the parameters $J_{ij}$.
We implement the algorithm in a MATLAB environment, with the minFunc routine for the L-BFGS function optimization \cite{schmidt2005minfunc}.  
To test the model and the inference procedure we choose a squared matrix having a side length of $w = 12$ px, both for the input and output signal, turning into $N=2w^2=288$ total intensity values. For this system, $\mathbb T$ is a $w^2 \times w^2$ matrix with  $w^4=20736$ coupling parameters to be estimated. On the other side, we are optimizing the cost function of the system, Eq. (\ref{eq:hamiltonian}) via the coupling matrix $\mathbb J$, that now has $k=(2w^2)^2 = 82944$  parameters (not all independent).
We generate data to analyze through transmission matrices $\mathbb T$ built with a random activation of $20\%$  of its elements (set equal to $1$).  The value of each element is, then, rescaled by the number of total active parameters per row, so that $\sum_{\gamma=1}^{N/2} T_{\alpha\gamma} = 1$. In this way, selecting a random input intensity distribution in the range of $I_\alpha^{in} \in [0, 1]$, also gives $I_\gamma^{out} \in [0,1]$. We stress that, however, the results presented in the following are general, with no restriction for intensity range nor matrix dimension. To the output we add a Gaussian noise of null mean and mean square displacement in the range of $\sigma=[0, 0.5]$, running independent optimization per each $\sigma$. 
With this procedure we create a sample of $M=5000$ couples of input and outputs for the inference.
%, about one order of magnitude less than the number of free parameters.  

The green curve in Fig. \ref{fig:PSLsurf} represents the first minimization of Eq.  (\ref{eq:totalPSL}), with all the parameters free to variate,  without exploiting their relationships. Having a direct look at the so-inferred matrix $\mathbb J$, we find close matching with what expected from its tensor form Eq. (\ref{eq:generalizedJ}) by the knowledge of the ad hoc generated $\mathbb T$.

Decimating the smallest couplings keeps the $\mathcal{L}$ constant down to a certain point, where the curve abruptly decreases. Rather than using the tilted pseudolikelihood function \cite{Decelle2014}, we use the Bayesian Information Criterion (BIC) as in \cite{rocchi2019} used here, instead, to estimate the best decimated model:
\begin{align}
{\rm BIC} = k\ln (M) - 2\mathcal{PL}.
\end{align}
The best number of couplings minimize the BIC and is represented by the light blue points in the 3D plot of Fig. \ref{fig:PSLsurf} versus the number of decimated couplings and the noise, compared against the true number of active couplings (red points). We found good agreement for the parameters estimation up to a $\sigma=0.25$, after which the BIC estimation favors networks with smaller numbers of couplings with respect to the true one.
To test the faithfulness of the inferred $\mathbb T_{\rm inf}$, we calculate the reconstruction quality with respect the true $\mathbb T$ as $Q = \left({{\norm{\mathbb T- \mathbb T_{\rm inf}}}/{\norm{\mathbb{T}}}}\right)^{1/2}$.
%\begin{equation}
%Q = \sqrt{\frac{\norm{\mathbb T- \mathbb T_{\rm inf}}}{\norm{\mathbb{T}}}}.
%\end{equation}
Here, $Q=0$ correspond to exact recovery of $\mathbb{T_{\rm inf}}$. In Fig.\ref{fig:errorFocusing} we plot the error on a network whose couplings are selected by BIC by blue points and the error on the true network (though values of the non-zero inferred elements can vary) with a red line. The two curves follow the same trend up to a channel noise of $25\%$, beyond which the BIC error becomes systematically larger. However, when using $\mathbb T_{\rm inf}$ in a focusing experiment to transmit a Gaussian function the quality of the focusing rapidly drops already around $\sigma=0.08$.
% We believe this value to be strictly connected with the sparsity of $\mathbb T$, and robustness can be tested in further studies.
% {\bf{\color{red}{ Che vuoi dire qui esattamente? Qual'\`e il messaggio che vuoi trasmettere?}}}

\begin{figure}[b!]
\centerline{ \includegraphics[width=8cm]{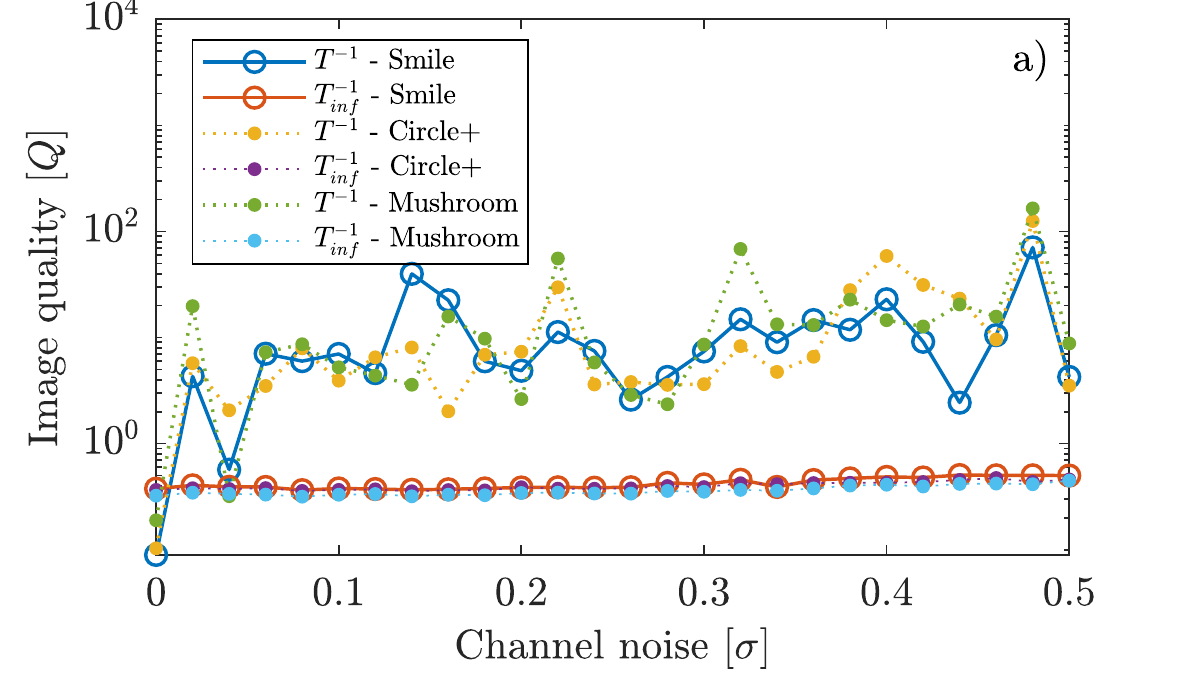}}
\centerline{ \adjincludegraphics[width=8cm,trim={0 {.3\height} 0 0 },clip]{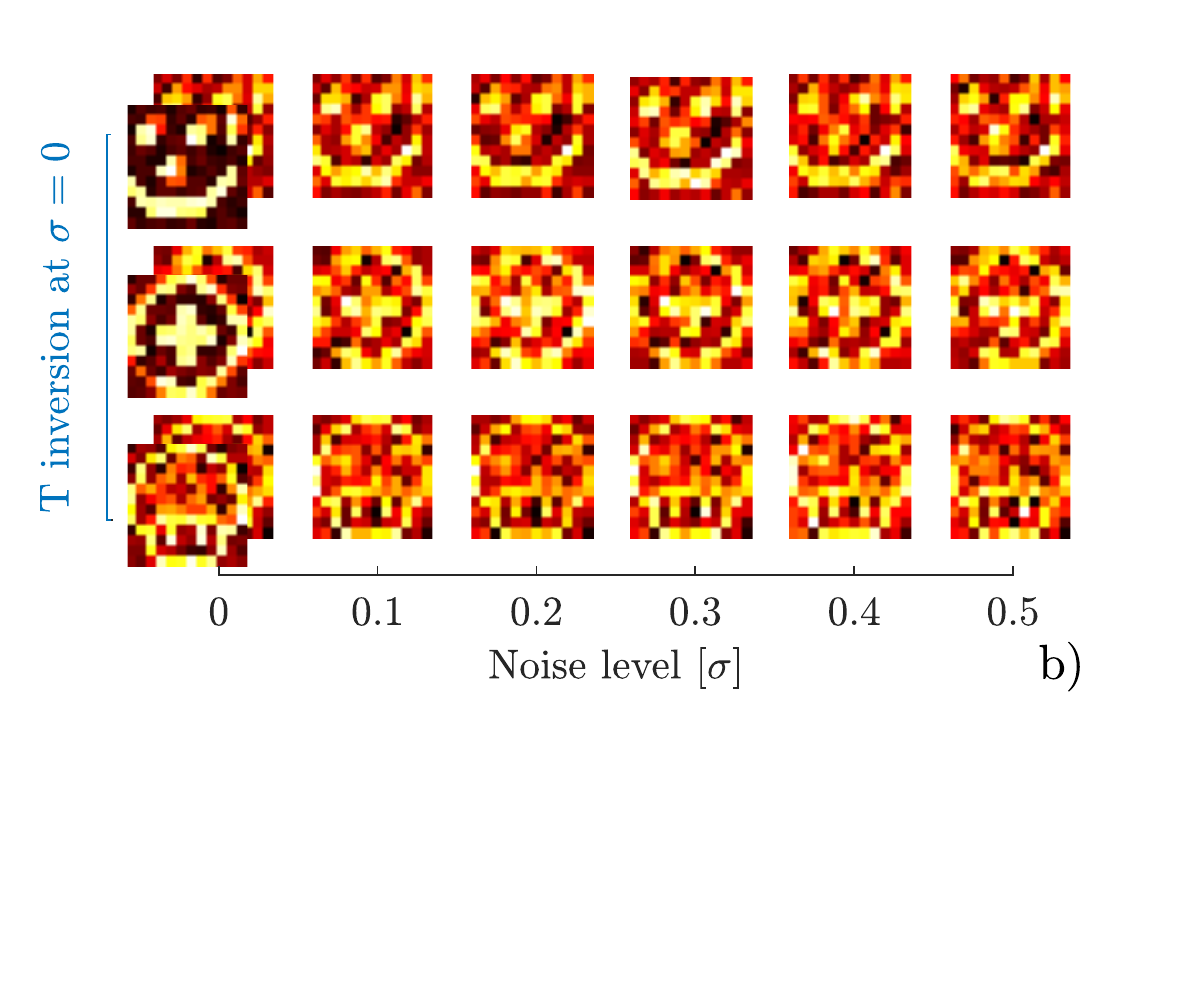}}
% \centerline{ \includegraphics[width=8cm]{latest_figures/Fig4_12x12_draws} }
\caption{Image reconstruction quality using the inferred $\mathbb T^{-1}$ in log-scale. It is possible to notice how the inversion of the direct $\mathbb T$ immediately degrades the reconstruction quality at low noise level, while the inference of the inverse transmission is highly stable up to the whole noise range studied.}
\label{fig:imagereconstruction}
\end{figure}

Our machine learning framework is symmetric under inversion, thus with the same inference protocol it is possible to obtain a valid reconstruction for the inverse transmission matrix $\mathbb T^{-1}$. We use the reversed intensity vector $\bm{ \bar I} =\{ I_{\alpha}^{\rm out}, I_{\gamma}^{\rm in}\}$, obtained swapping the input with the output. We run the same algorithm switching the roles of input and output in Eq. (\ref{eq:transmission}), using decimation and the BIC as before to locate the best inference point. At difference with the direct transmission matrix, that is sparse, the inverse matrix is not: with BIC we pick matrices having around $70\%$ active parameters.
% (most probably due to strong undersampling rate and broad noise-range explored). 
We test the efficiency of the $\mathbb T^{-1}$-recovery via an inverse image reconstruction process, where we send an object {{in input}} and we use the $\mathbb T^{-1}_{\rm inf}$ to recover the object from the speckled intensity distribution in output. 
Similarly with the previous definition, we define the image quality as the difference between the image sent $O$ and the reconstructed $O_{rec}$, thus $Q = \left( {{\norm{O-O_{rec}}}/{\norm{O}}} \right)^{1/2}$.
%\begin{equation}
%Q = \sqrt{\frac{\norm{O-O_{rec}}}{\norm{O}}}.
%\end{equation}
The method results robust in the whole perturbation range explored obtaining a uniform image quality as shown in Fig.\ref{fig:imagereconstruction}, part a. We can also observe that it is a much better performance with respect to the inverse of the inferred $\mathbb T$, as soon as the noise is non-zero. In Fig.\ref{fig:imagereconstruction}, part b we show how the object reconstructed is recognizable by the eye practically for any noise. Compared against the results of the direct $\mathbb T$ calculation, we observe that denser matrices  seem more robust to noise in all cases considered, though dedicated studies are required before making a general statement out of this evidence. 

{{Finally, it is important to notice that $\beta_{i> N/2}$ plays the role of an inverse temperature in the inference process. It is the effective noise variance per each channel, that in the output part turns into $\beta_\gamma \rightarrow 1/2/\sigma_\gamma$, where we now consider an explicit dependence of the noise on the channel, cf. Eq. \eqref{eq:delta_lim}  (see also Ref. [\onlinecite{Ancora19b}] for details). This information is automatically inferred in the diagonal part of $\mathbb J$ and serves to rule out $\mathbb T$ and $\mathbb U$, giving information about the transmission efficiency through the disordered channel.}}

%%%%%%%%%%%%%%%% SECTION
%\section{\label{sec:conclusions} Conclusions}
The model developed let us estimate I/O intensity coupling matrices via a machine learning approach and several benefits emerge from the usage of pseudolikelihood formulation. First of all, the pseudolikelihood can be calculated -and minimized- per each $i$-th element, making the computation scalable and trivially parallel, linearly depending on hardware resources (such as numbers of CUDA cores or independent GPUs). Moreover, the model is directly applicable for the estimation of the inverse coupling matrix $\mathbb T^{-1}$, rather than using matrix inversion or time reversal approaches \cite{Popoff2010natcomm,Carpenter2014}, experimentally extremely sensitive to noise. Due to its self consistent nature, the model gives us two further important information, such as the noise-estimation per channel (given by the diagonal part of the output-output coupling $\mathbb J$) and the balance criteria in the input-input Gramian matrix $\mathbb U$, which can be used to monitor algorithm convergence and as halt criteria. Lastly, our procedure can be generalized to any I/O system in any linear-regression scenarios. Here we considered intensities to provide a protocol ready to process experimental data, but the model can be extended straightforwardly to complex fields \cite{Popoff2010,Tyagi2016}, and it is possible to add polarization, i. e., vectorial waves, and wavelength dependence of $\mathbb T$ elements as further (linearly scalable) degrees of freedom. In the case of complex fields, the problem of the phase measurements is still present, but we could overcome it by integrating out the partition function over all the possible phases. In this case the pseudolikelihood formulation is more complicated, but the approach would be identical. In fact, rather than focusing on the exact model, this letter wants to introduce a first statistical framework for the $\mathbb T$-estimation. The usage of statistical models let us interpret the $\mathbb T$-recovery problem in terms of the minimization of the entropy of a system described by the coupling $\mathbb J$, thus into a thermodinamical problem, borrowing a number of tools coming from statistical mechanics. Thus, it opens up numerous opportunities, such as the study on the influence of the noise in the inference process up to study possible presence of phase transitions in terms of $\mathbb T$-matrix sparsity. This could lead to important characterization of structured focusing patterns \cite{DiBattista2016,di2016tailored} via its intrinsic transmission properties, or to study Anderson localization in 2D optical disordered systems \cite{Leonetti2014,Ruocco2017}, besides offering a statistical framework to study light propagation through opaque media.

%\begin{figure*}
%\subfloat{ \includegraphics[height=6cm]{Fig3a_plot1} }
%\subfloat{ \includegraphics[height=6cm]{Fig3b_inverseT} }
%\caption{Image reconstruction quality using the inferred $T^{-1}$ in log-scale. It is possible to notice how the inversion of the direct $T$ immediately degradate the reconstruction quality at low noise level, while the inference of the inverse transmission is highly stable up to the whole noise range studied.}
%\label{}
%\end{figure*}

\medskip
\bibliographystyle{unsrt} %Used BibTeX style is unsrt
\bibliography{bibliography_PRL}

\end{document}